\documentclass[10pt,twocolumn,letterpaper]{article}

\usepackage[pagenumbers]{cvpr} 

\renewcommand{\paragraph}[1]{\vspace{.5em}\noindent\textbf{#1}}

\definecolor{lgg}{rgb}{0.75, 0.95, 0.65} 
\usepackage{multirow}

\definecolor{cvprblue}{rgb}{0.21,0.49,0.74}
\usepackage[pagebackref,breaklinks,colorlinks,allcolors=cvprblue]{hyperref}
\usepackage{footmisc}
\title{Reconstruction as a Bridge for Event-Based Visual Question Answering}

\author{Hanyue Lou$^{\#1,2}$~~Jiayi Zhou$^{\#1}$~~Yang Zhang$^{1}$~~Boyu Li$^{1}$~~Yi Wang$^{3}$~~Guangnan Ye$^{4,2}$~~Boxin Shi$^{1*}$\\
{\small $^1$~Peking University}~
{\small $^2$~Shanghai Innovation Institute} 
~
{\small $^3$~Shanghai AI Laboratory}
~
{\small $^4$ Fudan University}\\
\small\texttt{\{hylz,~liboyu,~shiboxin\}@pku.edu.cn}~~~
\small\texttt{\{flyfeather,~zhangyang2004\}@stu.pku.edu.cn}\\
\small\texttt{wangyi@pjlab.org.cn~~yegn@fudan.edu.cn}
}

\begin{document}

\twocolumn[{
\renewcommand\twocolumn[1][]{#1}

\maketitle

}]

\let\thefootnote=\relax
\footnotetext{$^1$~Hanyue Lou, Boyu Li, and Boxin Shi are with the State Key Laboratory of Multimedia Information Processing and the National Engineering Research Center of Visual Technology, School of Computer Science, Peking University.~~ $^{\#}$~Equal contribution.~~$^*$~Corresponding author.}
\footnotetext{$^{\$}$~~Code \& dataset: \url{https://github.com/HYLZ-2019/EvQA}}

\begin{abstract}
Integrating event cameras with Multimodal Large Language Models (MLLMs) promises general scene understanding in challenging visual conditions, yet requires navigating a trade-off between preserving the unique advantages of event data and ensuring compatibility with frame-based models. 
We address this challenge by using reconstruction as a bridge, proposing a straightforward Frame-based Reconstruction and Tokenization (FRT) method and designing an efficient Adaptive Reconstruction and Tokenization (ART) method that leverages event sparsity.
For robust evaluation, we introduce EvQA, the first objective, real-world benchmark for event-based MLLMs, comprising 1,000 event-Q\&A pairs from 22 public datasets. Our experiments demonstrate that our methods achieve state-of-the-art performance on EvQA, highlighting the significant potential of MLLMs in event-based vision.
\end{abstract}    
\section{Introduction}
\label{sec:intro}

\begin{table*}[!t]
	\centering
	\caption{Comparison of existing event-based MLLM benchmarks, with preferred properties highlighted in \colorbox{lgg}{green}. Our EvQA benchmark covers a wider range of datasets, utilizes real event data, and employs objective evaluation metrics.}
	\label{tab:related_benchmarks}
	\resizebox{\textwidth}{!}{
	\begin{tabular}{c c c c}
		\toprule
		\textbf{Dataset Name} & \textbf{Source Dataset Diversity} & \textbf{Event Fidelity} & \textbf{Objectivity} \\
		\midrule
		N-ImageNet-Chat~\cite{eventgpt} & N-ImageNet~\cite{N-ImageNet} & Semi-real & Subjective (GPT scoring) \\
		Event-Chat~\cite{eventgpt} & DSEC~\cite{gehrig2021dsec}, IJRR~\cite{mueggler2017ijrr} & \colorbox{lgg}{Real} & Subjective (GPT scoring) \\
		\midrule
		EventVL-QA~\cite{EventVL} & N-ImageNet, DSEC, HARDVS~\cite{HARDVS} & Semi-real+Real & Subjective (GPT scoring) \\
		EventVL-Ds.~\cite{EventVL} & N-ImageNet, N-Caltech101~\cite{orchard2015nmnist}, DSEC, HARDVS & Semi-real+Real & Subjective (Similarity metrics) \\
		\midrule
		EVQA-Bench~\cite{LET-US} & Video QA datasets + v2e~\cite{v2e_Hu2021} & Synthetic & \colorbox{lgg}{Objective (Multiple-choice)} \\
		\midrule
		\textbf{EvQA (Ours)} & \colorbox{lgg}{22 Public Datasets} & \colorbox{lgg}{Real} & \colorbox{lgg}{Objective (Multiple-choice)} \\
		\bottomrule
	\end{tabular}%
	}
\end{table*}

Event cameras are bio-inspired sensors that capture per-pixel brightness changes asynchronously, unlike traditional frame-based cameras~\cite{event_survey}. This novel sensing paradigm offers significant advantages including microsecond-level temporal resolution, high dynamic range, and low power consumption, making event cameras particularly effective in challenging scenarios such as high-speed motion, extreme lighting conditions, and long-term monitoring.

\begin{figure}[t]
	\centering
	\includegraphics[width=\linewidth]{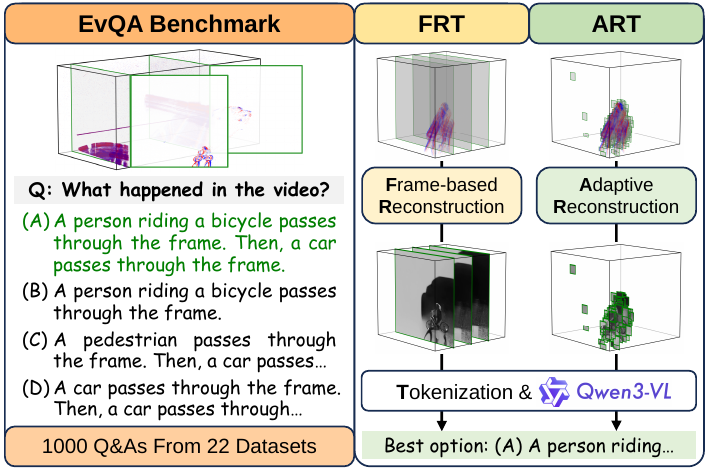}
	\caption{Left: We propose the EvQA benchmark for event-based MLLM question answering. Right: The FRT method prioritizes MLLM compatibility via frame-based reconstruction, while the ART method leverages the spatial sparsity of event streams with adaptive computation focusing on active spatio-temporal regions.}
	\label{fig:teaser}
\end{figure}

Researchers have developed many algorithms for various event-based vision tasks, including low level tasks such as deblurring~\cite{sun2023highrev} and high level tasks such as action classification~\cite{wang2024dailydvs}.
However, the application of event cameras to tasks requiring language abilities and high-level scene understanding, such as the question answering problem shown in the left side of ~\cref{fig:teaser}, remains largely unexplored.

In recent years, Multimodal Large Language Models (MLLMs) such as QwenVL~\cite{Qwen3VL}, InternVL~\cite{InternVL35}, Gemini~\cite{Gemini25}, and ChatGPT~\cite{GPT5} have demonstrated remarkable abilities in combining visual information with language instructions. However, these powerful models are designed for frame-based images and videos, and cannot directly handle the unique modality of event streams. This limitation raises a critical question: 
{\it How can we adapt MLLMs to event-based vision tasks?}

The core challenge lies in a fundamental trade-off. On one hand, MLLMs are optimized for standard images and videos; adapting them to novel modalities risks high computational costs for retraining and potential loss of valuable pre-trained knowledge. On the other hand, converting event streams into frame-based formats to fit MLLMs may discard their inherent advantages, such as high temporal resolution and spatial sparsity.

Existing methods like EventGPT~\cite{eventgpt}, EventVL~\cite{EventVL}, and LET-US~\cite{LET-US} attempt to bridge the modality gap by converting events into frame-based representations and then finetuning the MLLM backbone. These approaches, however, compromise on both fronts: they fail to fully leverage the unique properties of events and require costly model adaptation. 
In this paper, we explore reconstruction-based approaches that seek a more effective balance.

We first explore a method which prioritizes compatibility with existing MLLMs, Frame Reconstruction and Tokenization (FRT). This method reconstructs dense video frames from event streams using a state-of-the-art event-to-video model, V2V-E2VID~\cite{lou2025v2v}, and then feeds the videos into the Qwen3-VL~\cite{Qwen3VL} model. Our experiments show that this approach yields remarkable performance that scales with the base MLLM size and the reconstructed video frame rate, verifying that {\it reconstruction} can serve as a powerful bridge between event streams and MLLMs.

Based on the FRT method, we further design the Adaptive Reconstruction and Tokenization (ART) method, which aims to better exploit the sparsity of event streams. As compared to FRT in the right side of ~\cref{fig:teaser}, ART only triggers reconstruction in spatio-temporal regions with high event activity. 
To meet the unique requirements of this asynchronous paradigm, we employ mechanisms such as elapsed time embedding, selective state management and global feature exchangement to acquire a novel {\it Adaptive-E2VID} model, and modify the tokenization module of Qwen3-VL~\cite{Qwen3VL} to accommodate these sparse visual tokens.
Without finetuning any parameters of the MLLMs, the performance of ART also exceeds the prior work EventGPT~\cite{eventgpt} by a large margin. Compared to FRT, ART demonstrates significant reductions in token usage, especially on event sequences with high spatial sparsity.

In the emerging field of event-based MLLMs, the lack of an objective real-event benchmark hinders evaluation and comparison of different methods. We introduce EvQA, the first objective, real-event-based MLLM benchmark with high data diversity, addressing the limitations of existing benchmarks (Table~\ref{tab:related_benchmarks}). EvQA contains 1000 real-world event sequences from 22 public datasets, each with a manually annotated, objective multiple-choice question. The benchmark spans diverse scenarios, from street traffic to Antarctic wildlife, and includes data from 11 different event camera models. All questions are provided in both English and Chinese and have been validated by human experts. Our extensive experiments on EvQA confirm the effectiveness of our proposed methods.

In summary, our contributions are threefold:

\begin{itemize}
	\item We introduce EvQA, the first objective real-world benchmark for event-based MLLMs, built from 1000 diverse sequences across 22 datasets.
	\item We propose the FRT method and verify that reconstruction serves as a powerful bridge between event streams and MLLMs.
	\item We further design the ART method, which successfully leverages the sparsity of events for efficient MLLM processing.
\end{itemize}

\section{Related Works}
\label{sec:related}

\paragraph{Event-based MLLM Benchmarks.}
Several benchmarks for evaluating event-based MLLMs have been introduced by recent works~\cite{eventgpt, EventVL, LET-US}. However, these benchmarks face challenges regarding data fidelity, diversity, and evaluation objectivity.
\textbf{Data fidelity:} Due to the scarcity of real-world event datasets, many existing benchmarks rely on synthetic event data simulated from RGB videos or semi-real data captured by filming screens displaying static images~\cite{orchard2015nmnist, N-ImageNet}. This creates a domain gap between the evaluation setting and real-world applications.
\textbf{Data diversity:} Existing benchmarks often draw from a limited number of source datasets, failing to cover a wide range of real-world scenarios.
\textbf{Evaluation objectivity:} Many benchmarks focus on subjective tasks like captioning or open-ended question answering. These tasks typically rely on LLM-based scoring methods, which can introduce biases, generate hallucinatory responses, and exhibit inconsistent judgments~\cite{chen2024judge, wang2024notfair}.
As shown in Table~\ref{tab:related_benchmarks}, our proposed EvQA benchmark is the first to provide objective evaluation on real event data, with data diversity that surpasses all existing benchmarks.

\begin{figure*}[t]
	\centering
	\includegraphics[width=\textwidth]{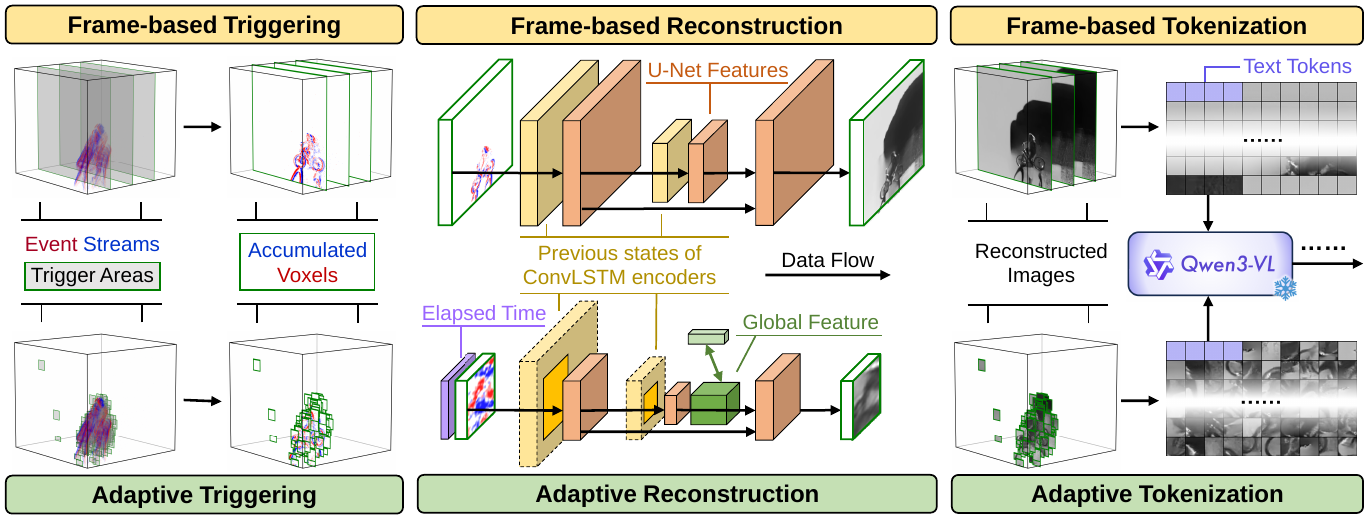}
	\caption{Our methods FRT and ART are both composed of three stages: accumulating events to voxels when triggered, reconstructing voxels to images, and tokenization for MLLM input. However, FRT employs frame-based strategies to maximize compatibility with MLLMs, while ART adaptively allocates computational resources based on event activity, efficiently leveraging the sparsity of event data.}
	\label{fig:pipeline}
\end{figure*}

\paragraph{Event-based MLLM Models.}
While pretrained vision models have been used to connect event data with language tasks in works like EventCLIP~\cite{wu2023eventclip} and EventBind~\cite{zhou2024eventbind}, the integration of events into MLLMs is a relatively new area.
EventGPT~\cite{eventgpt} represents the first attempt at this integration, using a CLIP-ViT-L~\cite{radford2021clip} for visual encoding and a finetuned Vicuna-7B-v1.5~\cite{vicuna2023} as the LLM backbone. However, its applicability is limited to short event streams of up to 0.1 seconds.
EventVL~\cite{EventVL} combines an event encoder with an InternVL2-2B~\cite{chen2024internvl25} backbone, aligning the modalities through contrastive learning.
LET-US~\cite{LET-US} employs SigLIP2~\cite{zhai2023sigmoid} and DINOv2~\cite{oquab2024dinov2} for feature extraction and Llama3.2-3B~\cite{meta2024llama32} as its backbone. It applies cross-modal guided filtering and temporal compression to the extracted features, enabling it to process long event streams exceeding 100 seconds.
Despite these advancements, all existing models still adhere to a frame-based paradigm. They process event streams in synchronous temporal bins across the entire spatial resolution, rather than dynamically allocating computation based on event activity. This prevents them from fully leveraging the asynchronous and sparse nature of event data.

\section{Method}
\label{sec:method}

This section details our reconstruction-based solutions for adapting MLLMs to event-based vision, as illustrated in \cref{fig:pipeline}. We first introduce Frame Reconstruction and Tokenization (FRT) (\cref{sec:frt}), a straightforward yet powerful approach that prioritizes compatibility with existing MLLMs. Building on this baseline, we then present Adaptive Reconstruction and Tokenization (ART) (\cref{sec:art}), which incorporates several innovations to efficiently leverage the inherent sparsity of event data.

\subsection{Frame Reconstruction and Tokenization}
\label{sec:frt}

Unlike traditional frame-based cameras that capture images at fixed intervals, event cameras asynchronously record pixel-level brightness changes. The output of an event camera is a stream of events, formulated as:
\begin{equation}
E = \{e_i = (x_i, y_i, t_i, p_i)\}_{i=1}^{N},
\end{equation}
where for each event $e_i$, $(x_i, y_i)$ are the pixel coordinates, $t_i$ is the timestamp, and $p_i \in \{+1, -1\}$ is the polarity of the brightness change.

However, existing MLLMs are designed to process standard images and videos. A straightforward approach is to convert the event stream into a sequence of video frames that can be directly fed into these models. We term this method Frame Reconstruction and Tokenization (FRT).

Many methods have been proposed for event-based video reconstruction. Among them, V2V-E2VID~\cite{lou2025v2v}, a retrained version of E2VID~\cite{rebecq2019e2vid}, demonstrates state-of-the-art performance. We therefore adopt it for our FRT pipeline.

With V2V-E2VID~\cite{lou2025v2v}, the event stream is first accumulated into a sequence of voxel grids $\{V_t\}_{t=1}^T$, as illustrated in the ``Frame-based Triggering'' part of \cref{fig:pipeline}. Each voxel grid $V_t$ spans a time interval of $\Delta t$ and has a shape of $(B, H, W)$, where $B=5$ is the number of temporal bins, and $(H, W)$ is the spatial resolution. The voxel values represent the sum of event polarities within each spatiotemporal bin:
\begin{equation}
	V_t(b, x, y) = \sum_{e_i: ~t_i \in [(t+\frac{b-1}{B})\Delta t, (t+\frac{b}{B})\Delta t], x_i=x, y_i=y} p_i.
\label{eq:voxel}
\end{equation}

The voxel grids are then recurrently fed into a reconstruction network to generate video frames. As shown in the ``Frame-based Reconstruction'' part of \cref{fig:pipeline}, this network employs a U-Net~\cite{ronneberger2015unet} architecture with ConvLSTM~\cite{shi2015convlstm} layers. We denote the reconstruction network as $\mathcal{R}$ and its hidden states at time $t$ as $S_t$. The reconstructed video frames $\{F_t\}_{t=1}^T$ are generated as follows:
\begin{equation}
	F_t, ~S_t = \mathcal{R}(V_t, ~S_{t-1}).
\end{equation}

The reconstructed video is then processed by the Qwen3-VL~\cite{Qwen3VL} tokenizer. This tokenizer merges every two consecutive frames and splits each merged frame into non-overlapping $32 \times 32$ patches. Each patch, along with its spatial position, is encoded into a visual token. Visual attention is then computed among the tokens within each frame.

The timestamp of each frame is encoded as text and inserted before the corresponding visual tokens. Text tokens from the input prompt are concatenated with the visual tokens. Attention is calculated across all tokens to generate the final answer. Thus, assuming each text timestamp uses $N_{\text{time}}$ tokens, the total number of input tokens for FRT is:
\begin{equation}
N_{\text{FRT}} = N_{\text{text}} + \frac{T}{2}\times(\frac{H}{32}\times \frac{W}{32}+N_{\text{time}}).
\end{equation}

A higher temporal resolution can be achieved by increasing $T$, which means reconstructing frames more frequently. However, this comes at the cost of an increased number of visual tokens.

This tokenization process is illustrated in the ``Frame-based Tokenization'' part of \cref{fig:pipeline}. As shown in the example, event activity is concentrated in a small area, yet redundant tokens are generated for all spatial locations, including blank regions or areas with reconstruction artifacts. This observation motivates our development of a more efficient, event-native approach.

\begin{figure*}[t]
	\centering
	\includegraphics[width=\textwidth]{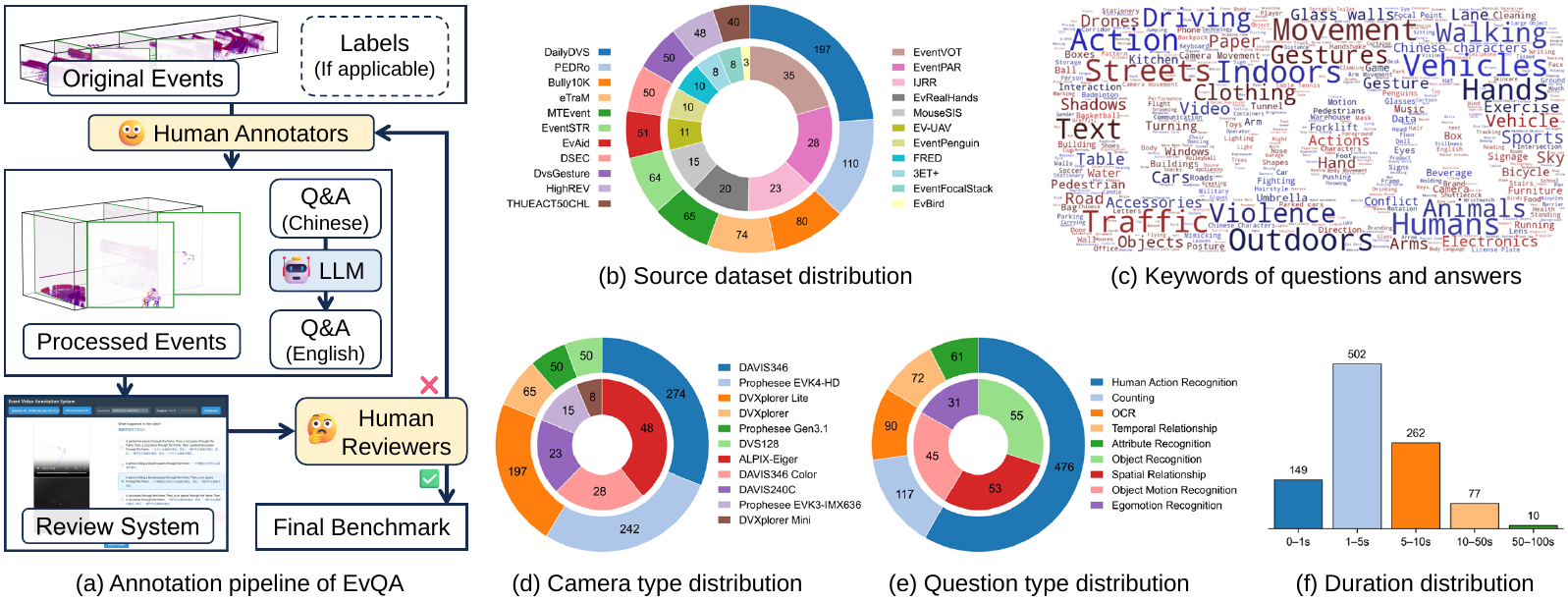}
	\caption{We construct EvQA, a real-event-based benchmark with human-annotated multiple choice questions. The event sequences are diverse over scenes, cameras and durations, while the questions cover a variety of types.}
	\label{fig:dataset}
\end{figure*}

\subsection{Adaptive Reconstruction and Tokenization}
\label{sec:art}
The dense, frame-based approach of FRT, while effective, fails to leverage the key advantages of event cameras: their asynchronous nature and data sparsity. To address this limitation, we propose Adaptive Reconstruction and Tokenization (ART), an event-native method that allocates computational resources based on event activity. ART modifies the standard reconstruction and tokenization pipeline in several key aspects.

\paragraph{Adaptive Triggering.}
Instead of using fixed time intervals, ART divides the scene into a grid of non-overlapping $32 \times 32$ patches and triggers reconstruction based on local event activity, as shown in the ``Adaptive Triggering'' part of \cref{fig:pipeline}. A patch is triggered for reconstruction only when the average number of new events per pixel within its boundaries exceeds a threshold $\theta$. In our experiments, we set $\theta = 0.5$. The events for the triggered patch are then accumulated into a local voxel grid, similar to \cref{eq:voxel}, but confined to the patch's spatiotemporal boundaries $(x_1, x_2, y_1, y_2, t_1, t_2)$, where $(x_1, y_1)$ and $(x_2, y_2)$ define the patch's spatial extent, and $t_1$ and $t_2$ are the timestamps of the last and current trigger events for that patch.

For efficiency, we process events in batches. Patches triggered by the same event batch are merged into larger rectangular regions using a greedy algorithm. Each resulting region, $r_j = (x_{j,1}, x_{j,2}, y_{j,1}, y_{j,2}, t_{j,2})$, is then processed by the reconstruction network in chronological order of its trigger time $t_{j,2}$. 
An overview of the proposed network, Adaptive-E2VID, is shown in the ``Adaptive Reconstruction'' part of \cref{fig:pipeline}.

\paragraph{Elapsed Time Embedding.}
The adaptive triggering mechanism results in non-uniform time intervals between reconstructions. To provide the model with this crucial temporal context, we introduce an ``elapsed time map'' as an additional input channel to the reconstruction network. For each pixel, this map stores the time elapsed since its last reconstruction, enabling the model to better understand the underlying temporal dynamics.

\paragraph{Selective State Management.}
The reconstruction network in ART maintains the full ConvLSTM hidden states for the entire spatial grid. However, during each reconstruction step for a region $r_j$, it selectively uses and updates only the hidden states corresponding to that region:
\begin{align}
	F_{j}, ~S_{j}(r_j) &= \mathcal{R}(V_{j}, ~S_{j-1}(r_j)),\\
	S_{j}(r_{\text{rest}}) &= S_{j-1}(r_{\text{rest}}),
\end{align}
where $r_{\text{rest}}$ denotes the spatial locations outside $r_j$. For deeper layers with lower spatial resolution, the corresponding active regions are determined by downsampling $r_j$. This strategy preserves long-term temporal memory across the entire scene while focusing computation only on areas with new information.

\paragraph{Global Feature Exchange.}
Processing patches in isolation can lead to a loss of global context. To address this, we introduce a global feature vector $f_g \in \mathbb{R}^{K}$ to facilitate information exchange across different patches. This vector interacts with the deepest feature map of each patch, which has $C$ channels and a spatial size of $(h, w)$.

When reconstructing a patch $r_j$, the previous global feature vector $f_{g}(j-1)$ is broadcast to match the patch's feature map size and concatenated, forming a combined feature map of shape $(C+K, h, w)$. This map is processed by a fusion layer, which outputs a refined feature map of the same shape. The first $C$ channels are passed to the next layer of the network, while the last $K$ channels, denoted $F_{g, \text{out}}$, are used to update the global feature vector:
\begin{equation}
\Delta f_g(j) = \text{AveragePool}(F_{g, \text{out}}),
\end{equation}
\begin{equation}
f_{g}(j) = f_g(j-1) + \Delta f_g(j).
\end{equation}

This mechanism is compatible between regions with different spatial shapes, which is required since the greedy merging process produces regions of varying sizes. It allows information to flow between regions of distinct positions and shapes, enhancing the model's ability to capture global context.

\paragraph{Adaptive Tokenization.}
Finally, we adapt the MLLM's tokenizer to handle the sparse and irregular stream of reconstructed patches. The tokenizer assigns the correct positional encoding to each patch based on its absolute spatial location within the full sensor resolution.

A key limitation of the Qwen3-VL~\cite{Qwen3VL} architecture lies in its handling of temporal information. Unlike Qwen2.5-VL~\cite{Qwen25VL}, which can assign a unique temporal encoding to each visual token, Qwen3-VL requires all tokens within the same conceptual frame to share a single, text-based timestamp. This prevents us from encoding the precise reconstruction time for each adaptively generated patch. To work around this, we group visual tokens into pseudo-frames of size $\text{TPF}$ (Tokens-Per-Frame) and insert a single timestamp token before each block. 
We set $\text{TPF} = 512$. 

To comply with Qwen3-VL's requirement of processing pairs of temporally adjacent patches from the same spatial location, our adaptive triggering mechanism is configured to always trigger patches in pairs.

For an event sequence that triggers $P$ patches, the total number of input tokens for ART is:
\begin{equation}
N_{\text{ART}} = N_{\text{text}} + \frac{P}{2} \times (1 + \frac{N_{\text{time}}}{\text{TPF}}),
\end{equation}
This number is independent of the sequence's total duration and depends only on the amount of event activity. This allows ART to allocate computational resources dynamically, leading to significant efficiency gains in scenarios with sparse events.

The ``Adaptive Tokenization'' part of \cref{fig:pipeline} illustrates this process. Only patches with significant event activity are reconstructed and tokenized, resulting in a more efficient representation that focuses on informative regions and minimizes redundancy.
\section{The EvQA Benchmark}
\label{sec:dataset}

As compared in \cref{tab:related_benchmarks}, existing event-based MLLM benchmarks are limited in data diversity, event fidelity and question objectivity. In addition, none of them have yet been completely released to the public. To address these issues, we present EvQA, a real-event-based MLLM benchmark with high data diversity and objective mutliple-choice questions. In this section, we introduce the diverse soruces of the event streams, the annotation process for generating mutliple-choice questions and our quality standards. An overview of the dataset is illustrated in \cref{fig:dataset}.

\subsection{Diverse Data Sources}

To construct a high-quality and diverse dataset, we curated a collection of publicly available, real-world event camera datasets. We deliberately excluded synthetic or semi-real datasets, such as N-ImageNet~\cite{N-ImageNet} and N-Caltech101~\cite{orchard2015nmnist}, which are generated by saccading a screen. These datasets often fail to capture the authentic motion dynamics of real-world objects and can introduce visual artifacts.

Our final selection comprises 22 distinct public datasets: 3ET+~\cite{chen20233et, wang20243et, chen20253et}, Bully10K~\cite{dong2023bullying10k}, DailyDVS~\cite{wang2024dailydvs}, DSEC~\cite{gehrig2021dsec}, DvsGesture~\cite{amir2017low}, eTraM~\cite{verma2024etram}, EV-UAV~\cite{chen2025evuav}, EvAid~\cite{duan2025eventaid}, EvBird~\cite{lou2025v2v}, EventFocalStack~\cite{lou2023all}, EventPAR~\cite{wang2025eventpar}, EventPenguin~\cite{hamann2024penguin}, EventSTR~\cite{wang2025eventstr}, EventVOT~\cite{wang2024eventvot}, EvRealHands~\cite{jiang2024evhandpose}, FRED~\cite{magrini2025fred}, HighREV~\cite{sun2023highrev, sun2024unified}, IJRR~\cite{mueggler2017ijrr}, MouseSIS~\cite{hamann2024mousesis}, MTEvent~\cite{awasthi2025mtevent}, PEDRo~\cite{boretti2023pedro} and THUEACT50CHL~\cite{gao2023thueact}. These datasets were captured using 11 different event camera models: ALPIX-Eiger, DAVIS240C, DAVIS346, DAVIS346 Color, DVS128, DVXplorer, DVXplorer Lite, DVXplorer Mini, Prophesee EVK3-IMX636, Prophesee EVK4-HD, and Prophesee Gen3.1. 

As illustrated in \cref{fig:dataset}(b) and (d), our dataset features a wide distribution of data sources and camera models. This diversity is crucial for developing models that can generalize to a variety of real-world conditions.

To ensure the accessibility of our work on public platforms such as HuggingFace, we verified that all source dataset licenses permit redistribution. For those without explicit licenses, we obtained permission directly from the original authors. Further details on the source datasets are provided in the supplementary material.

\subsection{Manual Annotation Process}

To address challenges from inconsistent data formats and the inability of current MLLMs to process raw event streams, we established a manual annotation pipeline where the authors served as both annotators and reviewers, as shown in \cref{fig:dataset}(a). This ensured the creation of high-quality question-answer pairs.

\paragraph{Event Processing.} Annotators began by sampling and cropping event data from the curated datasets, then converting them into a unified H5 format. Most sequences are 1-10 seconds long (\cref{fig:dataset}(f)), the longest spanning 97 seconds. Existing labels from some source datasets were used to generate draft questions to aid the process.

\paragraph{Question Generation.} By viewing event visualizations and reconstructed videos, annotators created objective multiple-choice questions, each with four options and one correct answer. To ensure diversity, we classified questions into nine categories: Object Recognition, Attribute Recognition, Object Motion Recognition, Human Action Recognition, Egomotion Recognition, Spatial Relationship, Temporal Relationship, Counting and Optical Character Recognition (OCR). Questions were balanced across nine categories, with Human Action Recognition being the most frequent (47.6\%) as shown in \cref{fig:dataset}(e).

\paragraph{Quality Review.} All annotations were manually reviewed to ensure a human observer would agree with the answer. The questions were initially written in Chinese and then translated to English with LLM assistance. As a final step, we used Qwen3~\cite{2025qwen3} to batch-verify the equivalence between the Chinese and English versions, ensuring translation accuracy.

\vspace{.5em}\noindent
In order to assist the annotators and reviewers, we developed a review system based on Flask. For each question, it visualizes event data by showing both accumulation videos (visualizing events in red and blue) and grayscale reconstructed videos produced with V2V-E2VID~\cite{lou2025v2v}. A ``Bad Question'' option is provided with the choices so reviewers can easily flag problematic questions. The system also keeps track of the dataset statistics to help the annotators improve data diversity.

\subsection{Quality Standards}

The core reason we chose manual annotation is that automatically generated questions often suffer from various quality issues, including but not limited to:

\begin{itemize}
\item Answer inconsistency: MLLMs may provide questions whose answers are inconsistent in the video, such as asking about the position of a moving object.
\item Position ambiguity: MLLMs may use ambiguous terms when describing spatial positions. For example, it may ask whether an object is ``in the left'' or ``in the center'' when it is actually 40\% from the left side of the frame: it is unclear which option is correct. Also, they often fail to distinguish between the camera frame's left/right and the filmed person's left/right hand side.
\item Counting ambiguity: Scenes often include partially visible objects, which introduces ambiguity to counting questions: Should they be counted or not?
\item Label noise: Errors exist in the labels of original datasets. For example, for sequences under the class label ``Raise both hands'', some actors only raise one hand. This causes automatically generated answers to be erroneous.
\item Insufficient information: MLLMs may generate questions related to information not visible by the event camera, such as color or static objects that did not trigger events.
\end{itemize}

To overcome these problems, we 
cropped sequences to remove ambiguous regions, used unambiguous language, fixed errors inherited from the original dataset labels, and ensured that all questions are objective and answerable based on the event data. Through this rigorous process, we established high-quality standards for the EvQA dataset.
\section{Experiments}
\label{sec:experiments}

In this section, we present a comprehensive evaluation of our proposed methods on the EvQA benchmark. We detail our experimental setup in \cref{sec:implementation_details}, followed by the results and analysis in \cref{sec:results_analysis}.

\subsection{Implementation Details}
\label{sec:implementation_details}

\paragraph{Experimental Setup.}
All experiments for our FRT and ART methods were conducted using the Qwen3-VL-Thinking~\cite{Qwen3VL} series of models, specifically the 2B, 4B, 8B, and 32B parameter versions, all operating in BF16 precision. All MLLM inference is performed based on the HuggingFace Transformers library~\cite{wolf2020transformers}. Following the protocol of MVBench~\cite{li2024mvbench}, we append the string \texttt{<|im\_start|>assistant Best option:(} to the end of each question prompt, so MLLMs can be guided to output a parsable single character (A, B, C, or D).

\paragraph{FRT Method.}
Our FRT implementation is entirely zero-shot. We use the official, unmodified V2V-E2VID weights from their public repository~\cite{lou2025v2v} to reconstruct videos from event streams at 24 FPS. For experiments requiring lower frame rates (0.1, 1, 2, 4, and 8 FPS), we uniformly subsample frames from the 24 FPS video. When feeding the video to the MLLM, we include the instructional prompt: "This is a low quality black and white video reconstructed from event streams."

\paragraph{ART Method.}
The Adaptive-E2VID model used in ART was trained from scratch using PyTorch. We modified the V2V~\cite{lou2025v2v} framework to simulate adaptive triggering: we convert video frames into voxel representations, calculating the incremental event count for each patch over time. A patch is triggered for reconstruction when the number of events per pixel exceeds a threshold. We train the model with 2000 videos from the WebVid~\cite{bain2021webvid} dataset. Note that no real events are used during training.

To manage training efficiency, we adopted a multi-stage curriculum with a batch size of 1 and a fixed learning rate of 1e-4. First, we pre-trained the model on full-frame (128$\times$128) reconstruction for 50 epochs. We then fine-tuned it for 5 epochs with a minimum patch size of 64$\times$64, followed by a final 5 epochs of fine-tuning with a 32$\times$32 minimum patch size. For the loss function, we use the L1 loss combined with a VGG version of the LPIPS loss.

For MLLM inference, the reconstructed sparse patches were accompanied by the prompt: ``We have reconstructed a low quality black and white video from event streams, here are its key patches (not complete) in chronological order.''

\subsection{Results and Analysis}
\label{sec:results_analysis}

To evaluate the performance of our methods, we use the accuracy (\%) on EvQA as the primary metric. We report results using Qwen3-VL models of varying sizes (2B, 4B, 8B, and 32B parameters), and track the average number of input tokens used during inference to assess computational efficiency. In addition to results on the full EvQA benchmark, we also present results on the EvQA-Sparse subset, which contains 200 sequences with lower event density, to highlight the efficiency advantages of our ART method. In the result tables, the highest accuracies and lowest token usages are highlighted in \colorbox{lgg}{green}.

\paragraph{Text-Only Baseline.}
To measure the guessability of the questions without visual input, we made the Qwen3-VL models guess with the prompt: ``The following question is about a lost video. Based on knowledge and reasoning, guess the most likely answer and select the best option from the provided choices.'' As shown in \cref{tab:text_only_results}, the accuracy is consistently above the 25\% random chance level but not very high, similar across the English and Chinese versions.

\begin{table}[t]
\centering
\caption{Text-only guessing accuracy (\%) on EvQA.}
\label{tab:text_only_results}
{\small
\begin{tabular}{lccccr}
\toprule
\textbf{Language} & \textbf{2B} & \textbf{4B} & \textbf{8B} & \textbf{32B} & \textbf{Tokens} \\
\midrule
English & 30.1 & 32.8 & 31.0 & 32.1 & 113.87 \\
Chinese & 31.3 & 31.3 & 31.3 & 35.4 & 110.83 \\
\bottomrule
\end{tabular}
}
\end{table}

\paragraph{FRT Results.} A key parameter for the FRT method is the frame rate (FPS) of the reconstructed video. The larger the frame rate, the more temporal information is preserved, but the number of tokens fed into the MLLM also increases. We experiment with frame rates of 0.1, 1, 2, 4, 8, and 24 FPS to analyze this trade-off.

As shown in \cref{tab:main_results}, the accuracy of FRT generally increases with both higher frame rates and larger model sizes. The best performance is achieved with the Qwen3-VL-32B model at 24 FPS, reaching an accuracy of 76.1\%. However, this comes at the cost of a high token count (14,798 tokens on average), making the method computationally expensive, especially for longer sequences.

\begin{table}[t]
\centering
\caption{Accuracy (\%) and average token usage of FRT and ART on the EvQA benchmark, scaling with model size.}
\label{tab:main_results}
{\small
\begin{tabular}{llccccr}
\toprule
\textbf{Method} & \textbf{FPS} & \textbf{2B} & \textbf{4B} & \textbf{8B} & \textbf{32B} & \textbf{Tokens} \\
\specialrule{\heavyrulewidth}{\aboverulesep}{\belowrulesep}
\multicolumn{7}{c}{Results on EvQA-Full (1000 Questions)} \\
\midrule
\multirow{6}{*}{FRT} & 0.1 & 55.1 & 56.6 & 58.8 & 63.2 & \colorbox{lgg}{738} \\
 & 1 & 56.8 & 58.3 & 61.5 & 66.1 & \colorbox{white}{963} \\
 & 2 & 60.3 & 62.1 & 65.6 & 68.7 & \colorbox{white}{1496} \\
 & 4 & 61.9 & 64.9 & 67.8 & 73.0 & \colorbox{white}{2742} \\
 & 8 & 63.8 & 67.7 & 69.4 & 73.9 & \colorbox{white}{5400} \\
 & 24 & 67.3 & 69.2 & 72.0 & \colorbox{lgg}{76.1} & \colorbox{white}{14798} \\
\midrule
ART & - & 46.3 & 49.3 & 50.9 & 57.9 & \colorbox{white}{1256} \\
\specialrule{\heavyrulewidth}{\aboverulesep}{\belowrulesep}
\multicolumn{7}{c}{Results on EvQA-Sparse (200 Questions)} \\
\midrule
\multirow{6}{*}{FRT} & 0.1 & 42.5 & 45.5 & 45.5 & 52.5 & \colorbox{white}{614} \\
 & 1 & 46.5 & 46.0 & 52.0 & 56.5 & \colorbox{white}{1103} \\
 & 2 & 50.0 & 56.0 & 57.5 & 61.0 & \colorbox{white}{1889} \\
 & 4 & 55.5 & 56.5 & 61.0 & 64.5 & \colorbox{white}{3643} \\
 & 8 & 57.0 & 59.5 & 60.5 & 65.5 & \colorbox{white}{7233} \\
 & 24 & 65.0 & 63.5 & \colorbox{lgg}{66.0} & \colorbox{lgg}{66.0} & \colorbox{white}{18352} \\
 \midrule
ART & - & 40.5 & 32.5 & 36.0 & 47.5 & \colorbox{lgg}{348} \\
\bottomrule
\end{tabular}
}
\end{table}

\paragraph{ART Results.}
The EvQA benchmark contains both dense event streams, such as those captured from a moving camera, and sparse event streams, such as those filming static scenes with occasional motion. In order to better evaluate the efficiency advantages of the ART method, we created a subset of EvQA called EvQA-Sparse. This subset consists of the 200 sequences with the lowest event density (events per second per pixel).

The results of ART on EvQA-Full and EvQA-Sparse are also presented in \cref{tab:main_results}. Although ART does not reach the same accuracy levels as FRT, it allows for significant reductions in token usage, especially on the sparse subset. On EvQA-Sparse, the ART method with Qwen3-VL-32B achieves an accuracy of 47.5\% while using only 348 tokens on average, which is less than 2\% of the tokens used by FRT at 24 FPS. This demonstrates the potential of ART for efficient event-based vision-language tasks.

Surprisingly, we observe that on EvQA-Sparse, the smaller ART-2B model outperforms the larger ART-4B and ART-8B models. A similar trend is seen with FRT, where the 2B model outperforms the 4B model at 1 FPS and 24 FPS. We attribute this behavior to the complex dynamics of MLLMs, which awaits further exploration.

\paragraph{Experiment with Tokens-Per-Frame.}
A key parameter for ART is the Tokens-Per-Frame (TPF) setting, which controls how many tokens are grouped into each pseudo-frame. With a smaller TPF, more temporal information is encoded via text timestamps; with a larger TPF, more inner-frame attention can be computed. We conducted an ablation study on TPF using the Qwen3-VL-2B model, with results shown in \cref{tab:ablation_tpf}. We find that a TPF of 512 achieves the best accuracy, balancing the two factors effectively.

\begin{table}[t]
\centering
\caption{Results on the effect of Tokens-Per-Frame (TPF) in the ART method using Qwen3-VL-2B.}
\label{tab:ablation_tpf}
{\small
\begin{tabular}{lcccccc}
\toprule
\textbf{TPF} & \textbf{64} & \textbf{128} & \textbf{256} & \textbf{512} & \textbf{1024} & \textbf{2048} \\
\midrule
Acc (\%) & 42.6 & 44.9 & 45.6 & \colorbox{lgg}{46.3} & 46.0 & 44.8 \\
Tokens & 1390 & 1313 & 1274 & 1256 & 1247 & \colorbox{lgg}{1243} \\
\bottomrule
\end{tabular}
}
\end{table}

\paragraph{Comparison with EventGPT.}
We finally compare our methods with EventGPT~\cite{eventgpt}, the only existing event-based MLLM method with open source code and weights. As EventGPT can only process event streams up to 0.1 seconds, we tested it by truncating all sequences in the EvQA benchmark, which range in duration from 0.19 to 97 seconds, to their first 0.1 seconds.

We split the EvQA questions according to the event durations into three groups: short sequences ($<$0.5s), medium sequences (0.5-10s), and long sequences ($>$10s). The results in \cref{tab:duration_split} show that a 0.1 second ``glimpse'' is insufficient for answering most questions, leading to EventGPT's accuracy decreasing as the duration of the original sequence increases. Our methods, FRT and ART, significantly outperform EventGPT across all duration groups.

\begin{table}[t]
\centering
\caption{Accuracy (\%) on EvQA, split by sequence duration. EventGPT~\cite{eventgpt} is evaluated on sequences truncated to 0.1s.}
\label{tab:duration_split}
{\small
\begin{tabular}{lccccc}
\toprule
\textbf{Method} & \textbf{Size} & \textbf{Total} & \textbf{$<$0.5s} & \textbf{0.5-10s} & \textbf{$>$10s} \\
\midrule
EventGPT~\cite{eventgpt} & 7B & 31.5 & 46.7 & 31.8 & 22.9 \\
FRT (24 FPS) & 2B & \colorbox{lgg}{67.3} & \colorbox{lgg}{76.7} & \colorbox{lgg}{66.4} & \colorbox{lgg}{73.5} \\
ART & 2B & 46.3 & 53.3 & 46.6 & 41.0 \\
\bottomrule
\end{tabular}
}
\end{table}

\paragraph{Qualitative results.}
Although we only quantitatively evaluate our methods on multiple-choice question answering, our methods are also capable of handling open-ended questions and other vision-language tasks. A qualitative example with visual grounding is shown in \cref{fig:qualitative}, with more results provided in the supplementary material.

\begin{figure}[t]
	\centering
	\includegraphics[width=\linewidth]{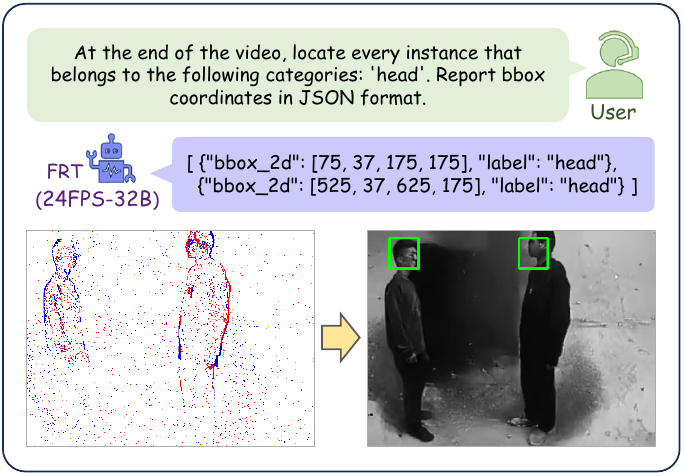}
	\caption{Qualitative results on a visual grounding task.}
	\label{fig:qualitative}
\end{figure}
\section{Conclusion}
\label{sec:conclusion}

In this paper, we explored the challenge of adapting MLLMs for event-based vision with reconstruction as a bridge. We introduced the approaches of FRT and ART, and contributed the first real-event-based objective MLLM benchmark, EvQA, composed of 1000 event sequences from 22 diverse datasets. Our experiments revealed that the straightforward FRT method achieves remarkable state-of-the-art performance, while ART serves as an important proof-of-concept for an efficient, sparsity-aware alternative. 

\paragraph{Limitations.}
While ART successfully leverages the sparsity of event streams, its departure from the conventional frame-based paradigm introduces significant challenges. On the reconstruction side, the dynamic shapes and positions of the generated patches in Adaptive-E2VID complicate efficient batching for parallel processing. On the MLLM side, existing models are only trained on frame-based visual data; adapting them to process sparse, ``shattered'' patches leads to performance degradation, as this format is out-of-distribution for their pre-trained mechanisms. Overcoming these hurdles will require future work on novel architectures that are natively designed and trained for sparse, asynchronous data.
\newpage
{
    \small
    \bibliographystyle{ieeenat_fullname}
    \bibliography{main}
}


\end{document}